\documentclass[12pt]{article}
\usepackage[a4paper, margin = 0.5in]{geometry}
\usepackage{times}
\usepackage{amsmath,amsfonts,amsthm}
\usepackage{graphicx}
\usepackage{caption}
\usepackage{subcaption}
\usepackage{abstract}
\usepackage{multicol}
\usepackage{multirow}

\title{\vspace{-6ex}Automatic Annotation of Axoplasmic Reticula in Pursuit of Connectomes}
\author {Ayushi Sinha \and William Gray Roncal \and Narayanan Kasthuri \and Ming Chuang \and Priya Manavalan \and Dean M. Kleissas \and Joshua T. Vogelstein \and R. Jacob Vogelstein \and Randal Burns \and Jeff W. Lichtman \and Michael Kazhdan}
\date{}
\begin{document}

%\twocolumn[
%\begin{@twocolumnfalse}

\maketitle
\thispagestyle{empty}
\pagestyle{empty}
\vspace{-9.5mm}
\begin{abstract}
\noindent
\textbf{\emph{Abstract}:} In this paper, we present a new pipeline which automatically identifies and annotates axoplasmic reticula, which are small subcellular structures present only in axons. We run our algorithm on the Kasthuri11 dataset, which was color corrected using gradient-domain techniques to adjust contrast. We use a bilateral filter to smooth out the noise in this data while preserving edges, which highlights axoplasmic reticula. These axoplasmic reticula are then annotated using a morphological region growing algorithm. Additionally, we perform Laplacian sharpening on the bilaterally filtered data to enhance edges, and repeat the morphological region growing algorithm to annotate more axoplasmic reticula. We track our annotations through the slices to improve precision, and to create long objects to aid in segment merging. This method annotates axoplasmic reticula with high precision. Our algorithm can easily be adapted to annotate axoplasmic reticula in different sets of brain data by changing a few thresholds. The contribution of this work is the introduction of a straightforward and robust pipeline which annotates axoplasmic reticula with high precision, contributing towards advancements in automatic feature annotations in neural EM data.\\
\end{abstract}
%\end{@twocolumnfalse}
%]

\begin{multicols}{2}
\section{Introduction}The Open Connectome project (located at http://openconnecto.me) aims to annotate all the features in a 3D volume of neural EM data, connect these features, and compute a high resolution wiring diagram of the brain, known as a connectome. It is hoped that such work will help elucidate the structure and function of the human brain.

The aim of this work is to automatically annotate axoplasmic reticula, since it is extremely time consuming to hand-annotate them. Specifically, the objective is to achieve an operating point with high precision, to enable robust contextual inference.  There has been very little previous work towards this end \cite{Kevin}. Axoplasmic reticula are present only in axons, indicating the identity of the surrounding process and informing automatic segmentation.

\section{Procedure}The brain data we are working with was color corrected using gradient-domain image-stitching techniques \cite{MH08} to adjust contrast through the slices. We use this data as the testbed for running our filters and annotating axoplasmic reticula.

\subsection{Image Processing}
\subsubsection{Bilateral Filter}The bilateral filter \cite{SPJF08} is a non-linear filter consisting of one 2D Gaussian kernel $G_{\sigma_{s}}$, which decays with spatial distance, and one 1D Gaussian kernel $G_{\sigma_{r}}$, which decays with pixel intensity:
\begin{equation*}
\begin{split}
	&\hspace{2mm}B[I]_p =  \frac{1}{W_p}\sum_{q\in S}G_{\sigma_{s}}(||p-q||)G_{\sigma_{r}}(I_p-I_q)I_q,\\
	&\hspace{4mm}\textrm{where }W_p = \sum_{q\in S}G_{\sigma_{s}}(||p-q||)G_{\sigma_{r}}(I_p-I_q)
\end{split}
\end{equation*}
is the normalization factor. This filter smooths the data by averaging over neighboring pixels while preserving edges, and consequently important detail, by not averaging over pixels with large intensity difference. Applying this filter accentuates features like axoplasmic reticula in our data.

\subsubsection{Laplacian Sharpening}Even with a narrow Gaussian in the intensity domain, the bilateral filter causes some color bleeding across edges. We try to undo this effect through Laplacian sharpening. The Laplacian filter computes the difference between the intensity at a pixel and the average intensity of its neighbors. Therefore, adding a Laplacian filtered image to the original image results in an increase in intensity where the average intensity of the surrounding pixels is less than that of the center pixel, an intensity drop where the average is greater, and no change in areas of constant intensity. Hence, we use the 3x3 Laplacian filter to highlight edges around dark features such as axoplasmic reticula. 

\vspace{3.3mm}
\subsection{Morphological Region Growing}We use a morphological region growing algorithm on our filtered data to locate and annotate axoplasmic

\end{multicols}
\begin{figure*}[h]
	\begin{subfigure}{26.5mm}
		\includegraphics[scale=0.43]{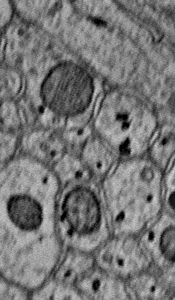}
		\caption{}
	\end{subfigure}
	$\Bigg\}$
	\begin{subfigure}{26.5mm}
		\includegraphics[scale=0.43]{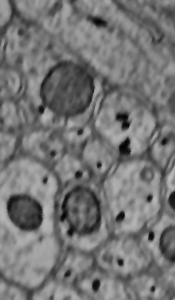}
		\caption{}
	\end{subfigure}
	$\Bigg\}$
	\begin{subfigure}{26.5mm}
		\includegraphics[scale=0.43]{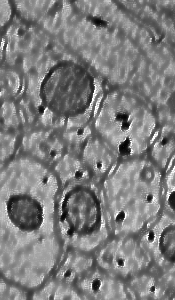}
		\caption{}
	\end{subfigure}
	$\Bigg\}$
	\begin{subfigure}{26.5mm}
		\includegraphics[scale=0.43]{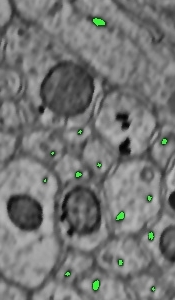}
		\caption{}
	\end{subfigure}
	$\Bigg\}$
	\begin{subfigure}{26.5mm}
		\includegraphics[scale=0.43]{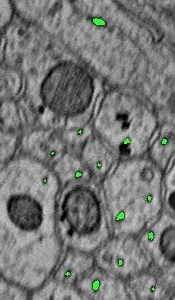}
		\caption{}
	\end{subfigure}
\hspace{3.75mm}
	\begin{subfigure}{26.5mm}
		\includegraphics[scale=0.43]{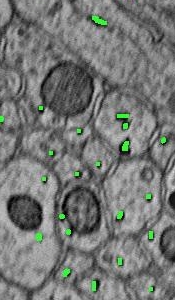}
		\caption{}
	\end{subfigure}
	\caption{Pipeline: (a) Original (color corrected) data, (b) Bilaterally filtered data, (c) Laplacian filtered data, (d) Annotations after performing morphological region growing on the bilaterally filtered data and the Laplacian sharpened data, (e) Annotations after tracking; and (f) Ground truth.}
\end{figure*}

\begin{multicols}{2}
\noindent reticula. We implement this by iterating over the filtered image and looking for dark pixels, where a dark pixel is defined as a pixel with value less than a certain specified threshold. When a dark pixel is found, we check its 8-neighborhood to determine if the surrounding pixels are also below the threshold. Then, we check the pixels surrounding these, and we do this until we find only high intensity pixels, or until we grow larger than the diameter of an axoplasmic reticula. The thresholds we use in our algorithm are biologically motivated and tuned empirically.

\subsection{Tracking}Finally, we track our annotations through the volume to verify their correctness and identify axoplasmic reticula that were missed initially. For each slice, we traverse the annotations and check if an axoplasmic reticulum is present in the corresponding xy-location (with some tolerance) in either of the adjacent slices. If a previously annotated axoplasmic reticulum object is present, we confirm the existing annotation.  Otherwise, the adjacent slice locations are checked for axoplasmic reticula with a less restrictive growing algorithm, and new annotations are added in the corresponding slice.  If no axoplasmic reticulum object is found in either of the adjacent slices, then we assume the annotation in the current slice to be incorrect, and delete it. 

\section{Results and Future Work} We qualitatively evaluated our algorithm on 20 slices from the Kasthuri11 dataset, and quantitatively compared our results against ground truth from a neurobiologist. Our algorithm annotates axoplasmic reticulum objects with 87\% precision, and 52\% recall.  These numbers are approximate since there is inherent ambiguity even among expert annotators.
\begin{center}
\begin{tabular}{c|c|c|}
\multicolumn{1}{c}{}
 &  \multicolumn{1}{c}{Actual Class}
 &  \multicolumn{1}{c}{(Observation)}\\
\cline{2-3}
		 	& 117: true  	& 18: false\\
Predicted Class	& positives  	& positives\\
\cline{2-3}
(Expectation) 	& 109: false 	& N/A : true\\
			& negatives 	& negatives\\
\cline{2-3}
\end{tabular}
\end{center}
\begin{center}
Table 1: Confusion matrix
\end{center}

Our current algorithm is designed to detect transverally sliced axoplasmic reticula. In future work, we plan to extend our morphological region growing algorithm to also find dilated axoplasmic reticula, and to incorporate a more robust tracking method such as Kalman or particle filtering. Additionally, our algorithm can be adapted to annotate other features in neural EM data, such as mitochondria, by modifying the morphological region growing algorithm.

%\section{Acknowledgments}This work is supported in part by grant number 1RO1EB016411-01 from the National Institute of Biomedical Imaging and Bioengineering at the National Institutes of Health.

\end{multicols}
\end{document}